\documentclass[sigconf,natbib=false]{acmart}

\usepackage{graphicx}
\usepackage{amsmath}
\usepackage{multirow}
\usepackage{caption}
\usepackage{subcaption}
\usepackage{lipsum}
\newcommand{\method}{CoMoDaL~}

\AtBeginDocument{%
  }

\renewcommand\footnotetextcopyrightpermission[1]{}

\copyrightyear{2023}
\acmYear{2023}
\setcopyright{acmlicensed}
\acmConference[MM '23] {Proceedings of the 31st ACM International Conference on Multimedia}{October 29--November 3, 2023}{Ottawa, ON, Canada.}
\acmBooktitle{Proceedings of the 31st ACM International Conference on Multimedia (MM '23), October 29--November 3, 2023, Ottawa, ON, Canada}
\acmPrice{15.00}
\acmISBN{979-8-4007-0108-5/23/10}
\acmDOI{10.1145/3581783.3612320}
\acmSubmissionID{2747}
\RequirePackage[
  datamodel=acmdatamodel,
  style=acmnumeric,
  ]{biblatex}

\begin{document}

\title{Cross-modal \& Cross-domain  Learning  for Unsupervised LiDAR Semantic Segmentation}

\author{Yiyang Chen}
\affiliation{%
  \institution{South China University of Technology}
  \city{Guangzhou}
  \country{China}
  }
\email{eeyychen@mail.scut.edu.cn}
\author{Shanshan Zhao}
\authornote{Corresponding authors}
\affiliation{%
  \institution{JD Explore Academy}
  \city{Beijing}
  \country{China}}
\email{sshan.zhao00@gmail.com}
\author{Changxing Ding}
\authornotemark[1]
\affiliation{%
  \institution{South China University of Technology}
  \institution{Pazhou Lab}
  \city{Guangzhou}
  \country{China}}
\email{chxding@scut.edu.cn}
\author{Liyao Tang}
\affiliation{%
  \institution{The University of Sydney}
  \city{Sydney}
  \country{Australia}}
\email{ltan9687@uni.sydney.edu.au}
\author{Chaoyue Wang}
\affiliation{%
  \institution{JD Explore Academy}
  \city{Beijing}
  \country{China}}
\email{chaoyue.wang@outlook.com}
\author{Dacheng Tao}
\affiliation{%
  \institution{The University of Sydney}
  \city{Sydney}
  \country{Australia}}
\email{dacheng.tao@gmail.com}

\renewcommand{\shortauthors}{Yiyang Chen et al.}

\begin{abstract}
  In recent years, 
cross-modal
domain adaptation has been studied on the paired 2D image and 3D LiDAR data to ease the labeling costs for 3D LiDAR semantic segmentation (3DLSS) in the target domain.
However, in such a setting the paired 2D and 3D data in the source domain are still collected with additional effort. 
Since the 2D-3D projections can enable the 3D model to learn semantic information from the 2D counterpart, we ask whether we could further remove the need of source 3D data and only rely on the source 2D images. To answer it, this paper studies a new 3DLSS setting where a 2D dataset (source) with semantic annotations and a paired but unannotated 2D image and 3D LiDAR data (target) are available~\footnote{Here, we still use the terms `source' and `target' in domain adaptation problem for clear presentation.}.
To achieve 3DLSS in this scenario, we propose \textbf{C}r\textbf{o}ss-\textbf{Mo}dal and Cross-\textbf{D}om\textbf{a}in \textbf{L}earning (CoMoDaL). 
Specifically, our \method aims at modeling 
1) inter-modal cross-domain distillation between the unpaired source 2D image and target 3D LiDAR data, and 2) the intra-domain cross-modal guidance between the target 2D image and 3D LiDAR data pair.
In \method, we propose to apply several constraints, such as point-to-pixel and prototype-to-pixel alignments, to associate the semantics in different modalities and domains by constructing mixed samples in two modalities.
The experimental results on several datasets show that in the proposed setting, the developed \method can achieve segmentation without the supervision of labeled LiDAR data. Ablations are also conducted to provide more analysis. Code will be available publicly\footnote{https://github.com/wdttt/comodal\_3d}.

\end{abstract}


\maketitle

\section{Introduction}
    \begin{figure}[htbp]
  \centering
  \includegraphics[width=\linewidth]{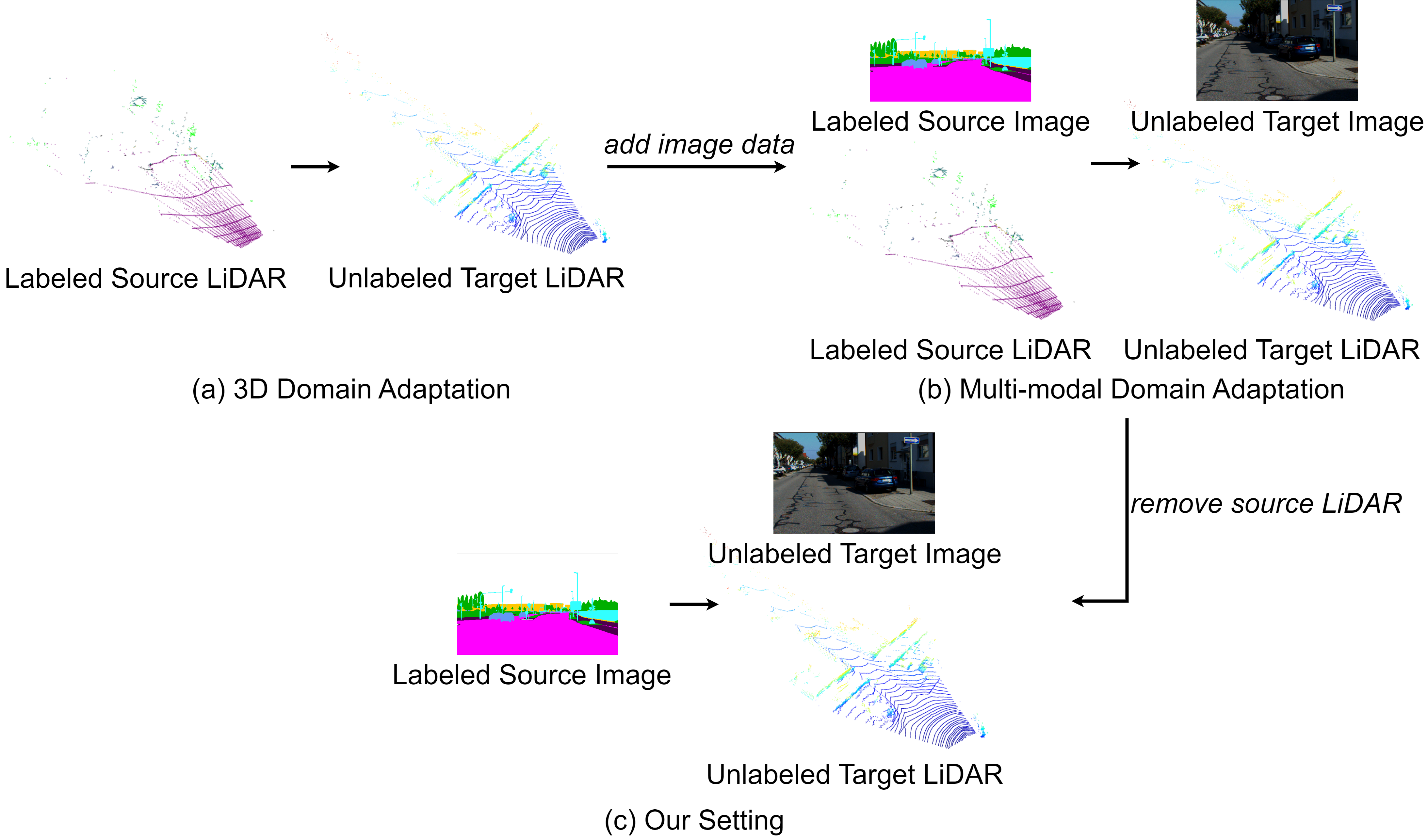} 
      \caption{Illustration of three different settings. (a) 3D Domain Adaptation: labeled 3D LiDAR data in source domain and unlabeled 3D LiDAR data in target domain; (b) Multi-modal Domain Adaptation: labeled image-LiDAR pair in source domain and unlabeled image-LiDAR pair in target domain; (c) Our Setting: labeled image in source domain and  unlabeled image-LiDAR pair in target domain. In all three settings, the main aim is to learn a 3D LiDAR semantic segmentation (3DLSS) model.
      } 
  \label{fig:setting} 
\end{figure}
{\color{black}
3DLSS is a fundamental vision task in the autonomous driving scenario~\cite{wu2018squeezeseg,zhou2020cylinder3d} and impressive performance has been achieved relying on large amounts of labeled 3D LiDAR data~\cite{nuscenes2019,behley2019iccv,geyer2020a2d2}.
Currently, domain adaptation (DA)~\cite{Yi_2021_CVPR,10.1007/978-3-031-19827-4_33,10.1007/978-3-031-19827-4_34} has been widely investigated on 3DLSS which seeks to remove the need on labels for a new dataset when a labeled source data is already available. For example, CosMix~\cite{10.1007/978-3-031-19827-4_34} mitigates the domain shift by introducing new training samples obtained by a sample mixing strategy.} 

{\color{black}
Apart from the uni-modal domain adaptation across different LiDAR data~\cite{Yi_2021_CVPR,10.1007/978-3-031-19827-4_33,10.1007/978-3-031-19827-4_34}, there are some methods exploring the DA between multi-modal data (2D-3D pair) for 3DLSS learning, such as xMUDA~\cite{jaritz2019xmuda} and DsCML~\cite{peng2021sparse}. They aim to mine the complementary advantages between 2D and 3D data by exploring the distillation between paired 2D-3D data.
Motivated by the study in xMUDA and DsCML, we wonder whether we could further remove the reliance on the 3D source data by exploring the 2D-3D projection and their complementarity. In this way, we can achieve 3D segmentation in the target domain requiring only labeled 2D images in our source domain. 
The comparison among different settings is shown in Fig.~\ref{fig:setting}. Since we do not have any labeled 3D information, how to provide the 3D network with effective supervision and construct the alignment between 2D and 3D is the issue to be solved in this paper.


Specifically, we propose cross-modal and cross-domain learning (CoMoDaL) to tightly associate the cross-modal and cross-domain data, \textit{i.e.,} 2D source image, 2D target image and 3D target LiDAR. CoMoDaL mainly consists of two modules: 1) inter-modal cross-domain distillation (ICD), and 2) intra-domain cross-modal guidance (ICG).
In detail, for ICD, we aim at aligning 2D source data and 3D target data. To make 2D source data paired with 3D data, we first introduce mixed images with CutMix strategy~\cite{9008296}, which not only contain pixels from the source images but also retain the correspondence with the 3D data. However, for the lack of 3D source data, point-to-pixel alignment cannot be applied to pixels from source images in the mixed images. In order to give an explicit constraint on these pixels for better regularization,
inspired by recent works based on class prototypes~\cite{10.1007/978-3-031-20056-4_3,Liu_2021_CVPR}, we devise a prototype-to-pixel alignment strategy. The developed prototype-to-pixel is able to achieve alignments by matching the pixel of the source 2D image to a prototype of the target 3D LiDAR. 
For ICG, we focus on the interaction between 2D target data and 3D target data. 
We enforce consistency between predictions of mixed 3D data and target 2D data to enhance the intra-domain point-pixel alignment. 
As the supervision from 2D network is noisy and unstable, we further introduce a 3D exponential moving average (EMA) teacher~\cite{NIPS2017_68053af2} to provide more stable pseudo-labels for the mixed LiDAR.

The main contributions of this paper are summarized as
follows:

\begin{itemize}
  \item We study a new setting, where 2D image data with dense semantic annotations in the source domain and paired image-LiDAR data without any semantic label in the target domain are available. 
  To achieve the 3DLSS in this setting, we propose a cross-modal and cross-domain learning strategy to fully mine the interactions between the two modalities.
  
  \item 
  In our CoMoDaL, we introduce two modules: ICD module can help build a correspondence for unpaired 2D source data and 3D target data to achieve more sufficient 2D-3D interaction, and ICG module applies the point-pixel alignment to the augmented 3D samples to enhance the intra-domain cross-modal learning.

  \item We conduct evaluation on the synthetic-to-real adaptation setting, GTA5-to-SemanticKITTI, and three real-to-real adaptation settings, including A2D2-to-SemanticKITTI, Cityscap-es-to-SemanticKITTI, and nuScenes Day-to-Night, to show the effectiveness of our method. We also evaluate previous related methods in our setting. 
\end{itemize}
\section{Related works}
    In this section, we review 
some previous methods for 3D  semantic segmentation, unsupervised domain adaptation for 3D segmentation, and sample mixing.
\newline

\noindent{\bf 3D  Semantic Segmentation.}
Currently, there are numerous point cloud segmentation methods, which can be briefly divided into three categories, \textit{i.e.,} projection-based, voxel-based, and point-based. 
Motivated by the remarkable achievement of deep learning on 2D image segmentation, some works ~\cite{wu2018squeezeseg,wu2019squeezesegv2,xu2021squeezesegv3} exploit the projection strategy, which projects the 3D point cloud into a 2D image plane and then takes advantage of the techniques of 2D semantic segmentation methods. To avoid the loss of geometric details caused by projection, processing data in the 3D space is studied extensively. One solution is voxelization. In detail,
the point cloud is voxelized into 3D grids so that 3D CNNs can be applied to process the regular 3D data \cite{riegler2017octnet,graham2015sparse,su2018splatnet}. 
Although efficient, the voxel-based methods still suffer from the loss of resolution to some extent.
In comparison,
point-based methods \cite{qi2017pointnet,qi2017pointnet++} directly work on the point cloud without voxelization.
Qi \textit{et al.}~\cite{qi2017pointnet} makes the first attempt to this strategy by proposing to apply shared MLP to each point. Following this seminal work, many variants are proposed~\cite{wang2019dynamic,thomas2019kpconv} to exploit the local geometric structure better.
Currently, for 3D LiDAR semantic segmentation, voxel-based methods have been dominant. For example,
SparseConvNet~\cite{3DSemanticSegmentationWithSubmanifoldSparseConvNet} represents 3D data as a set of sparse voxels and only applies 3D convolution to these voxels, Cylinder3D~\cite{zhou2020cylinder3d} designs a 3D point cloud representation, which suits for the varying sparsity of driving-scene LiDAR point cloud. 
Different from those methods that attempt to design efficient operations or network backbones, this paper is focused on achieving 3DLSS without labels in the target domain by studying a new setting.
\\
\newline
{\bf  Unsupervised domain adaptation for 3D segmentation.}
Recently, unsupervised domain adaptation (UDA) has attracted a lot
of attention in 3D semantic segmentation, which can be applied in the cases of real-to-real~\cite{yi2021complete,9341508} and synthetic-to-real~\cite{wu2019squeezesegv2,10.1007/978-3-031-19827-4_34} scenarios. Especially, Wu \textit{et al.}~\cite{wu2019squeezesegv2} propose to narrow the gap between real and synthetic 3D data with intensity rendering, geodesic correlation, alignment, and progressive domain calibration. Considering that sparse 3D point clouds are sampled from 3D surfaces, Yi \textit{et al.}~\cite{yi2021complete} propose to recover the underlying surfaces, and then process semantic segmentation on the completed 3D surfaces, transforming the domain adaptation task into a 3D surface completion task. Some self-learning methods~\cite{10.1007/978-3-031-19827-4_34,ding2022doda} based on sample mixing strategy also demonstrate their effectiveness.
Besides, multi-modal input data (i.e., 2D image + 3D LiDAR) has been exploited in some works~\cite{jaritz2019xmuda,peng2021sparse} to explore the effect of multi-modal information interaction for UDA in 3D segmentation. xMUDA~\cite{jaritz2019xmuda} mainly relies on pixel-to-point alignment between paired image-LiDAR data to achieve cross-modal learning. DsCML~\cite{peng2021sparse} adopts the sparse-to-dense feature matching strategy to utilize dense 2D features which do not correspond to 3D data in the paired image-LiDAR data. In our paper, we aim for removing the need for 3D source data 
\begin{figure*}[htbp]
  \centering
  \includegraphics[width=\linewidth]{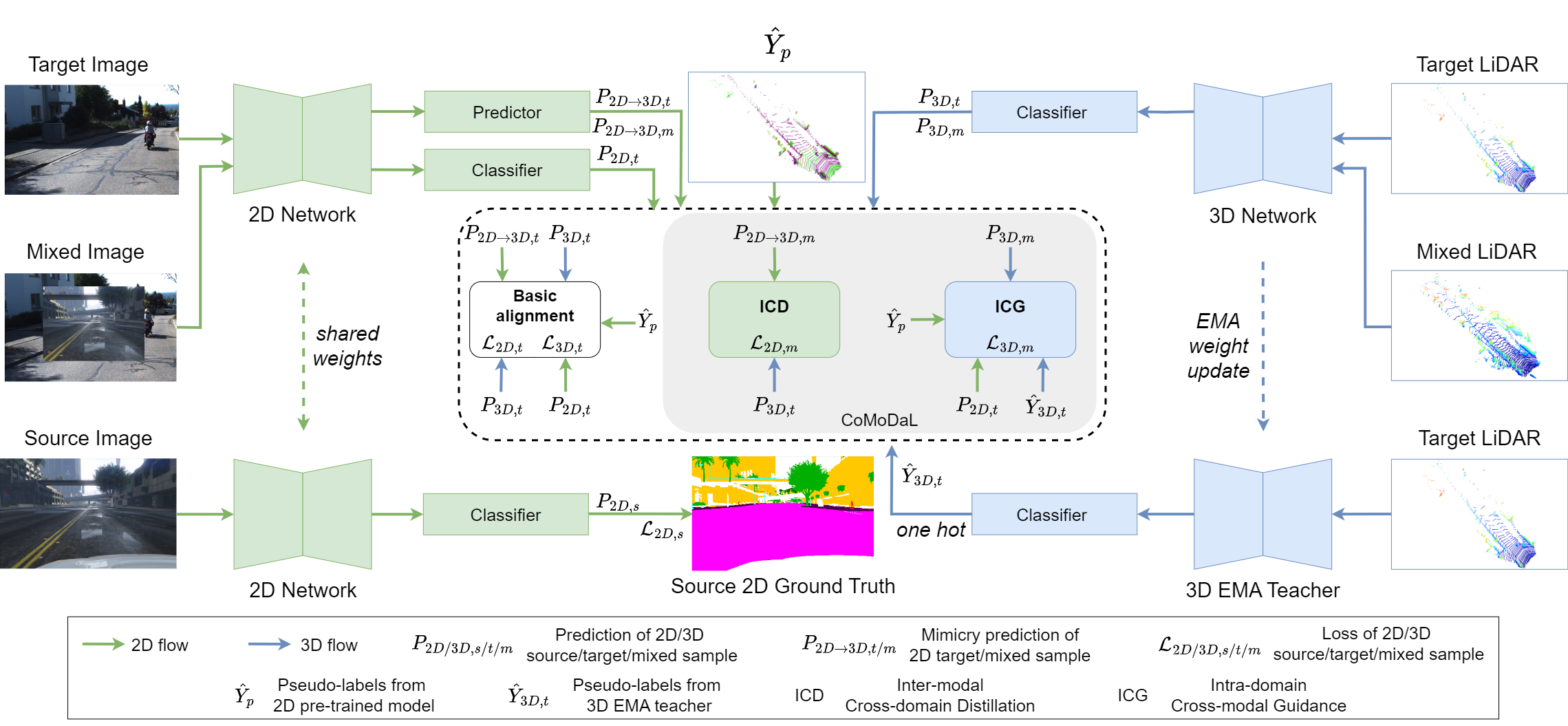} 
      \caption{Overall architecture of our approach. The predictor outputs mimicry prediction to estimate the other modality’s prediction while the classifier outputs prediction to achieve segmentation. Due to the lack of 3D data in the source domain, the basic alignment module can only tackle simple interactions between 2D and 3D in the target domain. Hence, we propose CoMoDaL to tightly associate the cross-modal and cross-domain data, which consists of two modules: ICD and ICG. ICD can promote the 2D-3D interaction via matching the mimicry prediction of 2D mixed sample ($P_{2D\rightarrow3D,m}$) to the prediction of 3D target sample ($P_{3D,t}$); ICG enforces consistency between the predictions of 3D mixed sample ($P_{3D,m}$) and 2D target sample ($P_{2D,t}$) to improve the basic intra-domain cross-modal learning, pseudo-labels from 3D exponential moving average (EMA) teacher ($\hat{Y}_{3D,t}$) are also introduced to make 3D network learn more stably.
      } 
  \label{fig:framwork} 
\end{figure*}
by fully exploring the interactions between the two modalities.
\\
\newline
{\bf Sample Mixing.}
Sample Mixing strategy has been widely used in self-supervised learning (SSL)~\cite{olsson2020classmix} and UDA~\cite{9423032,Hoyer_2021_CVPR,9710814,9879251} tasks, which introduces additional training data via either mixing the whole samples or cutting and pasting the patches to enhance the learning of samples without supervision. Classmix~\cite{olsson2020classmix} creates masks with the network's predictions for respecting object boundaries. In DACS~\cite{9423032}, mixed samples are created by mixing pairs of images from different domains. However, these methods rely on pixel-to-pixel correspondence between images to achieve and cannot be efficiently extended to sparse representations such as point clouds. 
For the 3D data, the sparsity and geometric structure need to be considered. For example,
PointMixup~\cite{chen2020pointmixup} realizes an optimal assignment of the path function between two point clouds to interpolate to create mixed samples. Mix3D~\cite{nekrasov2021mix3d} finds that directly concatenating two point clouds can make the model focus more on local structure rather than relying solely on contextual information. Lasermix~\cite{kong2023lasermix} mixes laser beams from different LiDAR scans to better leverage spatial prior of point cloud. 
This paper utilizes the mixing strategy to associate the data from different modalities and domains.
\section{Method}
    \subsection{Overview}
In this paper, we aim to learn a 3DLSS model by exploring a new setting where the 2D images in the source domain are labeled while the image-LiDAR pairs in the target domain are not.
We show the overview of our framework in Fig. \ref{fig:framwork}.
In our framework, there are two networks, one for the 3D LiDAR data (3DLSS) and one for the 2D image data (2D image segmentation). During training, the 2D network learns with the supervision of the ground truth segmentation map in the 2D source domain. 
In CoMoDaL, several alignments are developed to associate the different modalities and domains and to provide additional supervision for network learning. 
Exploiting the ground truth in the 2D source domain and the proposed CoMoDaL, we can achieve 3DLSS without any labeled 3D data.

Before detailing our method, we first give some used notations. The source dataset contains a set of 2D images $X_{2D,s}\in\mathbb{R}^{H\times W \times 3}$ with 2D segmentation labels $Y_{2D,s}\in\{1,2,...,C\}^{H\times W}$, while the target dataset contains 2D images $X_{2D,t}\in\mathbb{R}^{H\times W \times 3}$ and 3D point clouds $X_{3D,t}\in\mathbb{R}^{N \times 3}$ in the camera field of view, which have been synchronized and calibrated. Here, $H$, $W$, $C$, and $N$ denote the height, width, number of categories, and number of LiDAR points, respectively. Here, for brevity, we assume the images in the source and target domain have the same size ($H,W$).

\subsection{Basic alignment}
Previously, in both uni-modal LiDAR domain adaptation\cite{Yi_2021_CVPR,10.1007/978-3-031-19827-4_34} and multi-modal domain adaptation~\cite{jaritz2019xmuda,peng2021sparse}, there is labeled 3D data available and as a result, the 3D network can learn with point-wise semantic annotations. In contrast, in our setting, we have no access to the 3D labeled data, which causes challenges for 3D network learning. To alleviate the issue, we can use the predictions of the 2D network on the paired target images and sample them with the projection relationship from 3D to 2D to generate the pseudo-labels $\hat{Y}_{o}$:
\begin{equation}
    \hat{Y}_{o} =\mathop{\arg\max}\limits_{c\in  \{1,2,...,C\}}\Phi(P_{2D,t})^{(c)},
\end{equation}
where $P_{2D,t}$ is the prediction of 2D network on target images, the superscript $c$ indicates the prediction in the $c$th class, and $\Phi$ is the sampling function utilizing 3D-2D projection in target domain to sample 2D predictions. Then, the pseudo-labels  $\hat{Y}_{o}$ can be used to  supervise the 3D network.

However, such pseudo-labels have very low accuracy at the early training stages, which affects the learning of the 3D network. To provide a remedy, we propose a hybrid pseudo-labels strategy. Specifically, we first train a 2D segmentation model on the 2D source domain. The pre-trained 2D model can provide better pseudo-labels $\hat{Y}_{p}$ at the early training stages:
\begin{equation}
    \hat{Y}_{p} = \mathop{\arg\max}\limits_{c\in \{1,2,...,C\}} \Phi(\widetilde{P}_{2D,t})^{(c)},
\end{equation}
where $\Phi(\widetilde{P}_{2D,t})^{(c)}$ is the prediction  of 2D pre-trained model on sampled pixels in target images, and the superscript $c$ indicates the probability in the $c$th class. 
We propose to combine the predictions of the pre-trained 2D model and the 2D network trained with the 3D network together. We can achieve this using the Average or Max strategy. Here, we select the Max one to obtain hybrid pseudo-labels $\hat{Y}_{2D,t}$, which can be defined as follows:

\begin{equation}
\resizebox{0.9\linewidth}{!}{$
    \hat{Y}_{2D,t} = \left \{ \begin{aligned}
        &\hat{Y}_{o}, \quad  &\mathop{\max}\limits_{c\in  \{1,...,C\}} \Phi(\sigma(P_{2D,t}))^{(c)} \geq \mathop{\max}\limits_{c\in  \{1,...,C\}} \Phi(\sigma(\widetilde{P}_{2D,t}))^{(c)},\\
        &\hat{Y}_{p}, \quad  &\mathop{\max}\limits_{c\in  \{1,...,C\}} \Phi(\sigma(P_{2D,t}))^{(c)}<\mathop{\max}\limits_{c\in  \{1,...,C\}} \Phi(\sigma(\widetilde{P}_{2D,t}))^{(c)},
    \end{aligned}\right.$}
\end{equation}
where $\sigma$ represents the SoftMax function.
It means, we compare the output confidence of each model to select highly confident predictions to generate hybrid pseudo-labels for 2D data. Then, the 3D network can learn target samples with them:
\begin{equation}
    \mathcal{L}_{3D,t} = \mathcal{L}_{CE}(\sigma(P_{3D,t}),\hat{Y}_{2D,t}),
\end{equation}
where $P_{3D,t}$ is the prediction of the 3D network on target LiDAR, $\mathcal{L}_{CE}$ is the cross-entropy loss.

Furthermore,
since the supervision mainly comes from the 2D network, improving the performance of the 2D network would be beneficial for the 3D network to learn. Inspired  by xMUDA~\cite{jaritz2019xmuda}, we can enable the 2D network to learn from the 3D predictions in the target domain, which is implemented as follows: 
\begin{equation}
    \begin{aligned}
        \mathcal{L}_{2D,t} = \sum_{n=1}^{N}D_{KL}(\Phi(\sigma(P_{2D\rightarrow3D,t}))^{(n)}|| \sigma(P_{3D,t})^{(n)}),
    \end{aligned}
\end{equation}
where $D_{KL}$ is KL divergence. $\Phi(\sigma(P_{2D\rightarrow3D,t}))^{(n)}$ is the mimicry prediction of the 2D network on $n$th sampled pixel in target images. Mimicry prediction learns from the other modality’s prediction via KL divergence, which is the core of information interaction between 2D and 3D modalities.

With a pre-trained 2D model, we can train the 2D and 3D segmentation networks by optimizing the above loss functions. Fig. ~\ref{fig:illustration} (a) illustrates the basic alignment module. However, this basic alignment neither addresses the domain gap issues within the 2D source and target images nor mines the interactions between the 2D and 3D data. This paper attempts to provide a remedy by the \method strategy, which is detailed in the following.

\subsection{Cross-modal and cross-domain learning}


The proposed \method aims at tightly associating cross-modal and cross-domain data and enhancing the learning of 3D network via inter-modal cross-domain distillation and intra-domain cross-modal guidance. Here, we introduce them separately.
\\

\noindent{\bf Inter-modal Cross-domain Distillation (ICD).}
\begin{figure}[htbp]
  \centering
  \includegraphics[width=\linewidth]{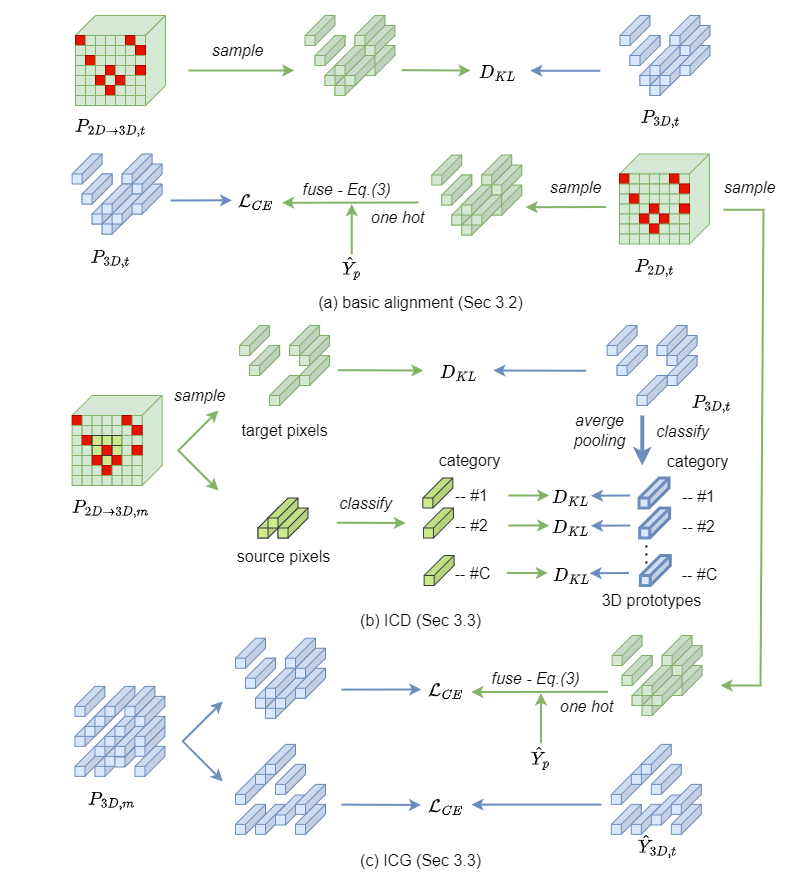} 
      \caption{Illustration of three modules. (a) In basic alignment, target images learn with 3D predictions while target LiDAR learns with hybrid pseudo-labels obtained from 2D network and 2D pre-trained model. (b) In ICD, for the mixed images, pixels from source images learn with corresponding 3D class prototypes while pixels from target images learn with corresponding 3D target predictions. (c) In ICG, mixed LiDAR learns with mixed pseudo-labels obtained from 2D network and 3D EMA teacher. }
  \label{fig:illustration} 
\end{figure}
To build a correspondence for 2D source data and 3D target data for more sufficient 2D-3D interaction, we propose to leverage CutMix strategy~\cite{9008296} to generate mixed samples to construct new image-LiDAR pairs so that the pixels from source images in the mixed images can be paired with 3D data. 

CutMix is an image augmentation technique that involves cutting out a patch from one image and pasting it onto another image, introducing more diverse training samples.
Formally, given a labeled source image $X_{2D,s}$ and an unlabeled target image $X_{2D,t}$,  the mixed image can be presented as:
\begin{equation}
    X_{2D,m} = M\odot X_{2D,s} + (1 - M)\odot X_{2D,t},
\end{equation}
where $M$ denotes a binary mask indicating which pixel needs to be copied from the source domain
and pasted to the target domain, $\odot$ represents the element-wise
multiplication operation, and $X_{2D,m}$ represents the mixed image. 

Although employing CutMix can make 2D source data paired with 3D target data, only the pixels from target images in the mixed images have a point-to-pixel correspondence with 3D target data. This kind of naive point-to-pixel alignment is limited to the strict point-wise image-LiDAR correspondence and is infeasible to be applied to the pixels from source images in the mixed images. Hence, it is nontrivial to construct an effective and explicit constraint that differs from point-to-pixel for better regularization. 

Some existing domain adaptation works~\cite{10.1007/978-3-031-20056-4_3,Liu_2021_CVPR} have explored how to build a correspondence for unpaired images from different domains based on class prototypes. 
Inspired by them, we design a novel prototype-to-pixel alignment to match unpaired 2D and 3D data. Specifically, we first calculate the class prototypes of 3D target predictions:
\begin{equation}
    \rho_{3D,t}^{(c)} = \frac{\sum_{n=1}^{N} P_{3D,t}^{(n)} * \mathbb{I}(\hat{Y}_{2D,t}^{(n)}==c)}{\sum_{n=1}^{N}  \mathbb{I}(\hat{Y}_{2D,t}^{(n)}==c)},
    \label{equation:7}
\end{equation}
where $\mathbb{I}$ is an indicator function returning true if the condition is satisfied. $\rho_{3D,t}^{(c)}$ is the 3D class prototype of class $c$. Eq. \ref{equation:7} represents that 3D class prototype $\rho_{3D,t}^{(c)}$ is obtained by averaging the predictions of points in target LiDAR belonging to class $c$. Due to the lack of annotations in the target domain, we use the pseudo-labels provided by the 2D network to classify the points in the target LiDAR.

With 2D source ground truth and 3D class prototypes, we determine a class prototype corresponding to which pixel from source images in the mixed images according to the ground truth. The 2D mixed predictions can learn from the following 3D predictions:

\begin{equation}
        \overline{P}_{3D,t}^{(n)} = \left \{ \begin{aligned}
        &\sum_{c=1}^{C}\mathbb{I}(\Phi{(Y_{2D,s})}^{(n)}==c)*\rho_{3D,t}^{(c)}, \quad  &\Phi(M)^{(n)}=1
        ,\\
        &P_{3D,t}^{(n)},\quad  &\Phi(M)^{(n)}=0,
    \end{aligned}\right.
    \label{equation:8}
\end{equation}
where $\Phi{(Y_{2D,s})}^{(n)}$ is the label of $n$th sampled pixel in source images, and $\Phi(M)^{(n)}$ is the value of $n$th sampled pixel in the binary masks.

Therefore, for the mixed images, the loss function can be written as: 
\begin{equation}
    \begin{aligned}
        \mathcal{L}_{2D,m} = \sum_{n=1}^{N}D_{KL}(\Phi(\sigma(P_{2D\rightarrow3D,m}))^{(n)}|| \sigma(\overline{P}_{3D,t})^{(n)}),
    \end{aligned}
\end{equation}
where $\Phi(\sigma(P_{2D\rightarrow3D,m}))^{(n)}$ is the mimicry prediction of the 2D network on $n$th sampled pixel in mixed images. It can be seen from Fig. ~\ref{fig:illustration} (b) that we actually apply prototype-to-pixel alignment for pixels from the source domain and point-to-pixel alignment for pixels from the target domain in mixed images. 
\\

\noindent{\bf Intra-domain Cross-modal Guidance (ICG).}
In the basic alignment, cross-modal learning has been applied between target images and LiDAR. Considering 3D branch lacks 3D source data and is only supervised by 2D branch in our setting, we employ data augmentation on 3D data and enforce consistency between predictions of augmented 3D data and target 2D data to enhance cross-modal learning. We adopt sample mixing introduced by Mix3D~\cite{nekrasov2021mix3d} as the augmentation strategy.

Mix3D introduces a mixing strategy to avoid strong contextual prior information impeding the model's learning of object semantics in 3D semantic segmentation task. This allows the model to avoid being over-reliant on strong contextual priors that may not always be reliable or relevant. 
Given a training batch $\{X_{3D,t}^{(1)},X_{3D,t}^{(2)},...,X_{3D,t}^{(B)}\}$, $B$ is the batch size, the mixed LiDAR can be presented as:
\begin{equation}
    X_{3D,m}^{(i)} = X_{3D,t}^{(i)} \cup X_{3D,t}^{(j)},
    \label{equation:10}
\end{equation}
where $i,j \in \{1,2...,B\}$, and $\cup$ is the merging operator. Eq. \ref{equation:10} represents the $i$th mixed LiDAR $X_{3D,m}^{(i)}$ in the new training batch is constructed by merging the $i$th LiDAR in the original training batch and any LiDAR within that batch.

However, Mix3D is realized under the condition that 3D data has ground truth as supervision. In contrast, the supervision of 3D data in our setting is noisy and unstable. 
Here we use Mean Teacher strategy~\cite{NIPS2017_68053af2} to generate more stable pseudo-labels to tackle this problem. We build a 3D EMA teacher model additionally, which updates its weights with the weights of student model (\textit{i.e.,} 3D network in training) via an exponential moving average algorithm as follows:
\begin{equation}
    \theta_{t} = \lambda\theta_{t} + (1-\lambda)\theta_{s},
\end{equation}
where $\theta_{s}$ represents the parameters of student model, $\theta_{t}$ represents the parameters of teacher model, and $\lambda$ is the hyperparameter to control the speed of update, which is set to 0.99.
The segmentation labels of 3D mixed samples $\hat{Y}_{3D,m}$ can be obtained as follows:
\begin{equation}
    \hat{Y}_{3D,m}^{(i)} = \hat{Y}_{2D,t}^{(i)} \cup \hat{Y}_{3D,t}^{(j)},
\end{equation}
where $\hat{Y}_{3D,t}^{(j)}$ is the pseudo-label of $j$th sample in the training batch obtained from the teacher model. Therefore, 3D network can learn mixed LiDAR as follows:
\begin{equation}
    \mathcal{L}_{3D,m} = \mathcal{L}_{CE}(\sigma(P_{3D,m}),\hat{Y}_{3D,m}),
\end{equation}
where $P_{3D,m}$ is the prediction of 3D network on mixed LiDAR. The illustration of ICG is shown in Fig. ~\ref{fig:illustration} (c). 

\subsection{Training objective}

The total loss on 2D branch can be written as:
\begin{equation}
    \begin{aligned}
        &\mathcal{L}_{2D}=  \mathcal{L}_{2D,s}
        +\lambda_{2D,t}\mathcal{L}_{2D,t} + \lambda_{2D,m}\mathcal{L}_{2D,m},
    \end{aligned}
\end{equation}
\begin{equation}
    \mathcal{L}_{2D,s} = \mathcal{L}_{CE}(\sigma(P_{2D,s}),Y_{2D,s}),
\end{equation}
where $\lambda_{2D,t},\lambda_{2D,m}$ are the hyperparameters to balance different losses, $P_{2D,s}$ is the prediction of 2D network on source images.

The total loss on 3D branch is:
\begin{equation}
    \mathcal{L}_{3D} = \mathcal{L}_{3D,t} + \lambda_{3D,m}\mathcal{L}_{3D,m},
\end{equation}
where $\lambda_{3D,m}$ is the hyperparameter to weight $\mathcal{L}_{3D,m}$.

\begin{table*}[t]
\begin{centering}

\begin{tabular}{ccccccccccccc}
\hline
\multirow{2}{*}{Method} & \multicolumn{3}{c}{A2D2-to-Sem.KITTI}                                 & \multicolumn{3}{c}{GTA5-to-Sem.KITTI}                                 & \multicolumn{3}{c}{Cityscapes-to-Sem.KITTI}
                            & \multicolumn{3}{c}{nuScenes Day-to-Night}\\ \cline{2-13} 
                        & 2D                   & 3D                   & Avg                  & 2D                   & 3D                   & Avg                  & 2D                   & 3D                   & Avg                  & 2D                   & 3D                   & Avg\\
\hline
Source                  &36.0                      & —                   & —                   &27.9                      & —                   & —                   &33.4                      & —                   & —                   &41.8                      & —                   & —                   \\
\hline
PL                      & 39.1          & 44.9         & 43.6         & 28.9          & 36.0         & 33.4         & 37.8            & 42.1           & 41.9           & 43.5            & 30.7           & 30.5      \\
MinEnt                  & 37.9          & 42.2         & 40.5         & 28.5       & 34.3         & 32.2         & 41.4         & 44.8           & 44.5          & 42.1            & 29.9           & 30.6 \\
DACS                  & 45.5          & 45.5         & 47.3         & 33.3       & 37.2         & 36.3         & 44.4        & 47.3          & 48.0          & \textcolor{red}{47.6}            & 34.2           & 36.7 \\
\hline
xMUDA*                   & 37.3                & 43.5                &41.9                      &33.9                      &39.0                      &38.0                      &44.6                      &43.6                      &49.4                      & 45.7            & 34.7           & 37.5 \\
xMUDA*+PL                 &39.7                      &46.8                      &43.2                      &34.1                      &40.3                      &41.1                      &46.0                      &45.8                      &51.0                      & 46.0            & 36.1           & \textcolor{red}{38.9} \\
DsCML*                   &39.9                      &45.2                      &43.4                      &31.8                      &37.5                      &37.4                      &43.0                      &44.3                      &47.8                      & 44.9            & 35.6           & 33.9 \\
DsCML*+PL                 &42.3                      &47.4                      &44.9                      &31.4                      &38.7                      &39.5                      &46.1                      &45.3                      &49.1                      & 45.2            & 36.0           & 34.9 \\
\hline
CoMoDaL                    & \textcolor{blue}{48.6}                      &\textcolor{blue}{48.2}                      &\textcolor{blue}{50.2}                      &\textcolor{blue}{42.2}                      &\textcolor{blue}{44.5}                      &\textcolor{blue}{45.6}                      &\textcolor{blue}{49.8}                      &\textcolor{blue}{49.6}                      &\textcolor{blue}{51.7}                     & \textcolor{blue}{47.5}            & \textcolor{blue}{37.0}           & \textcolor{blue}{37.9}\\
CoMoDaL+PL                  & \textcolor{red}{49.5}                      &\textcolor{red}{49.4}                      &\textcolor{red}{51.0}                      & \textcolor{red}{42.6}                     & \textcolor{red}{44.6}                      & \textcolor{red}{46.1}                      & \textcolor{red}{50.1}                      &\textcolor{red}{51.0}                      &\textcolor{red}{52.4}                      & \textcolor{red}{47.6}            & \textcolor{red}{37.1}           & 37.4\\
\hline
Oracle                 &58.3                      &71.0                     &73.7                      &55.6                      &63.6                      &67.1                      &55.6                      &63.6                      &67.1                      & 48.6            & 47.1           & 55.2\\
\hline
\end{tabular}
\caption{Comparison results of CoMoDaL and other adaptation methods for 3D semantic segmentation in four cross-modal domain adaptation settings, we report
the results for 2D and 3D streams as well as the ensembling result (‘softmax avg’). * indicates some modifications are made to adapt to our setting. The top-2 scores except the last row are marked by the \textcolor{red}{red} and \textcolor{blue}{blue}.
}
\label{tabel:main results}
\end{centering}
\end{table*}
\section{Experiment}
    \subsection{Datasets}
For fairly comparing with multi-modal domain adaptation works\cite{jaritz2019xmuda,peng2021sparse}, we follow xMUDA~\cite{jaritz2019xmuda} to evaluate our method on two adaptation settings: \textbf{A2D2-to-SemanticKITTI} and \textbf{nuScenes Day-to-Night}. In addition, we leverage existing
 2D datasets to introduce extra two adaptation settings: \textbf{GTA5-to-SemanticKITTI} and \textbf{Cityscapes-to-SemanticKITTI} to better demonstrate the effectiveness of our method.

A2D2~\cite{geyer2020a2d2} is an autonomous driving dataset containing 27,695 training images with pixel-wise semantic annotations, GTA5~\cite{Richter2016eccv} and Cityscapes~\cite{Cordts_2016_CVPR} are two popular 2D semantic segmentation datasets in the autonomous driving scene containing 24,966 and 2,975 training images, respectively. SemanticKITTI~\cite{behley2019iccv} consists of 18,029 training samples and 1,101 validation samples. In SemanticKITTI, only the front camera images are available, and therefore, following xMUDA~\cite{jaritz2019xmuda} and DsCML~\cite{peng2021sparse}, we only consider the front view image and the LiDAR points that are projected into it. nuScenes\cite{nuscenes2019} also provides both images and LiDAR, and we leverage nuScenes to generate the Day/Night split like xMUDA.
In addition to the nuScenes Day-to-Night case, since the categories of the target and source datasets in the other three cases are not consistent, we need a pre-process. 

Specifically, for A2D2-to-SemanticKITTI, we define 10 shared classes between the two datasets, including Car, Truck, Bike, Person, Road, Parking, Sidewalk, Building, Nature, and Other Objects; while for GTA5-to-SemanticKITTI, we define 12 shared classes between GTA5 and SemanticKITTI, \textit{i.e.,} Car, High-vehicle, Bike, Rider, Person, Road, Parking, Sidewalk, Building, vegetation, terrain, and Other Objects. The shared classes for Cityscapes-to-SemanticKITTI are the same as GTA5-to-SemanticKITTI. 

\subsection{Implementation Details}
\noindent{\bf Network Architecture.}
Since there is no prior work exploring the proposed setting, to better compare with the previous multi-modal domain adaptation works~\cite{jaritz2019xmuda,peng2021sparse} whose setting is similar to us, we follow xMUDA~\cite{jaritz2019xmuda} to construct the 2D network and 3D network. More specifically, we adopt a modified version of U-Net~\cite{ronneberger2015unet} with a ResNet34~\cite{7780459} encoder for the 2D network, while we use SparseConvNet~\cite{graham20173d} with U-Net architecture for the 3D network (downsampling 6-times). The voxel size is set to 5cm to guarantee there is one 3D point at most per voxel.

\noindent{\bf Training Strategy.} 
During the training period, 
we use the Adam optimizer~\cite{kingma2017adam} with $\beta_{1} = 0.9$ and $\beta_{2} = 0.999$ and initial learning rate of 0.001. The learning rate will be adjusted according to the poly learning policy~\cite{chen2017rethinking} with a poly power of 0.9. The batch size is set to 8 and the max training iteration is set to 100,000. 

\subsection{Main results}
We evaluate our approach on the above four cases and make comparisons with several previous uni-modal and multi-modal methods. For the uni-modal domain adaptation methods, we select pseudo-labeling (PL)~\cite{lee2013pseudo}, entropy minimization (MinEnt)~\cite{vu2019advent} and DACS~\cite{9423032}. We apply these uni-modal methods to 2D branch while using 2D branch to generate pseudo-labels for the 3D branch. For the multi-modal 
domain adaptation methods, we compare our approach with two typical methods, \textit{i.e.,} xMUDA~\cite{jaritz2019xmuda} and DsCML~\cite{peng2021sparse}. Note that, due to the differences in settings, we make some modifications to the way they are implemented. For example, in our re-implementation, the cross-modal learning loss in xMUDA and DsCML is only calculated in the target domain while not in the source for the lack of 3D source data. The Oracle means training with target samples only, except the Day/Night Oracle, where we used batches of 50\%/50\% Day/Night to avoid overfitting. The shared classes of A2D2-to-SemanticKITTI are different from GTA5-to-SemanticKITTI and Cityscapes-to-SemanticKITTI, which leads to different Oracle results between these adaptation settings.

Tab.~\ref{tabel:main results} shows the experimental results and comparisons. 
We report the mIoU of the target 2D image and 3D LiDAR. The `+PL' in the table means training from scratch again with pseudo-labels generated from the optimized model. In addition, following xMUDA and DsCML, we also report the result of `Avg', which means taking the mean of the probability predictions of 2D and 3D networks.
We can observe that CoMoDaL performs favorably against all methods in all four cases, bringing consistent improvement on 3D. 
It is worth noting that CoMoDaL even outperforms xMUDA*+PL and DsCML*+PL without further self-training in almost all of the metrics. 
Both quantitative and qualitative results demonstrate that although no any labeled 3D data is available in the proposed setting, CoMoDaL is able to yield higher performance compared with other methods.

Some visualized 3DLSS results on four adaptation settings are depicted in Fig. \ref{fig:qualitative results}. Example (a) shows that compared with xMUDA*+PL and DsCML*+PL, CoMoDaL+PL accurately classifies the “Sidewalk” and “Parking” which are easily confused. In example (b), CoMoDaL+PL succeeds in distinguishing the boundary between "Road" and "Sidewalk" while the other two fail. Both examples (c) and (d) demonstrate that CoMoDaL+PL is competent to predict some hard classes correctly.

\subsection{Ablation studies}
\begin{table}[t]
\begin{center}
\begin{tabular}{ccccccc}
\hline
\multirow{2}{*}{}   & \multirow{2}{*}{H.pl}     & \multirow{2}{*}{ICD} & \multirow{2}{*}{ICG} & \multicolumn{3}{c}{GTA5/Sem.KITTI} \\ \cline{5-7} 
                                             &                           &                           &                           & 2D        & 3D        & avg        \\
\hline
\#1                &                           &                           &                           & 33.9         & 39.0         & 38.0         \\
\#2                & \checkmark &                           &                           & 40.7         &42.0         & 43.5          \\
\#3                & \checkmark & \checkmark &                           & 42.1         & 43.4         & 44.4          \\
\#4               & \checkmark & \checkmark & \checkmark & 42.2        & 44.5         & 45.6        \\
\hline
\end{tabular}
\caption{Ablation study on the effect of different components. 
H.pl means hybrid pseudo-labels, Row \#1 represents the performance of xMUDA*, and Row \#2 represents the performance of the basic alignment module. } 
\label{tabel:ablation}

\end{center}
\end{table}
\begin{figure*}[ht]
  \centering
  \includegraphics[width=\linewidth]{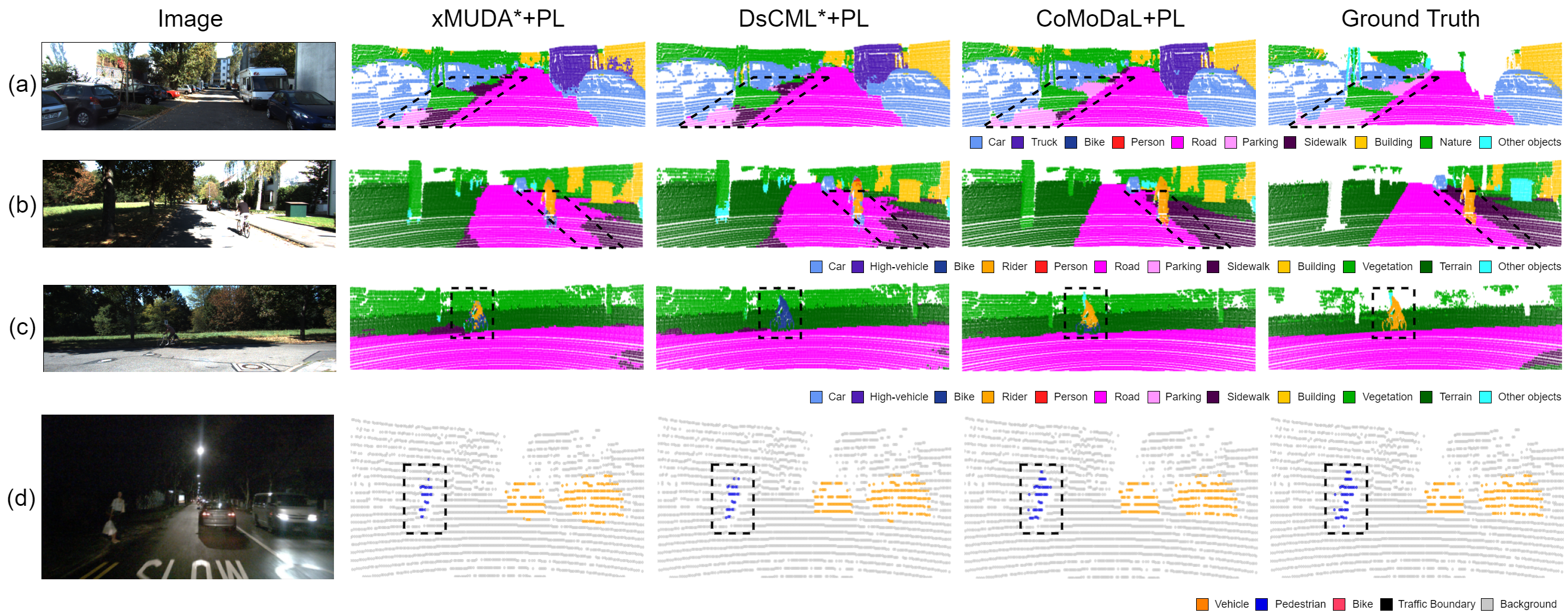} 
      \caption{Qualitative results on four adaptation settings: A2D2-to-SemanticKITTI, GTA5-to-SemanticKITTI, Cityscapes-to-SemanticKITTI, and nuScenes Day-to-Night, which are represented by (a), (b), (c) and (d) respectively in the figure. The visualized results are obtained by the output of 3D network. Compared to `xMUDA*+PL' and `DsCML*+PL', `CoMoDaL+PL' significantly improves the performance of segmentation.  } 
  \label{fig:qualitative results} 
\end{figure*}


\noindent{\bf Effects of different components.}
We carry out a series of experiments to investigate the developed CoMoDaL and report the performance in Tab.~\ref{tabel:ablation}. In Row \#1 we provide the performance of modified xMUDA that adapts to our setting, where 2D source data learns with ground truth and target image-LiDAR data pairs learn with each other via KL divergence. 
We first analyze the effect of hybrid pseudo-labels by comparing Row \#1 and Row \#2. Specifically,  1) In Row \#1 the 3D model learns  from the 2D model in training via KL divergence; 2) In Row \#2 the 3D model learns with the pseudo-labels from 2D pre-trained model and 2D model in training (\textit{i.e., }hybrid pseudo-labels). The comparison between Row \#1 and Row \#2 demonstrates that the in-training model and the pre-trained model can provide better supervision by performing together, which contributes to 6.8\% (2D), 3.0\% (3D) and 5.5\% (avg) mIOU gains.
The comparison between Row \#2 and Row \#3 indicates that ICD (inter-modal cross-domain distillation) enhances the performance of the 2D network by 1.4\%, the 3D network by 1.4\%, and the average by 0.9\%. The results show that ICD can contribute to promoting the interaction between 2D and 3D modalities.
Furthermore, from the comparison of Row \#3 and Row \#4, the inclusion of ICG (intra-domain cross-modal guidance) results in an improvement of 1.1\% for the 3D network, which confirms the role of ICG. 
\newline

\noindent{\bf Analyse of ICD.}
\begin{table}[ht]
\begin{center}
\begin{tabular}{cccccc}
\hline
\multicolumn{1}{l}{} & \multirow{2}{*}{mix} & \multirow{2}{*}{prototype} & \multicolumn{3}{c}{GTA5/Sem.KITTI} \\ \cline{4-6} 
\multicolumn{1}{l}{} &                      &                        & 2D     & 3D    & avg    \\
\hline
\#1                  & \checkmark                     &                        & 41.5      & 41.5     & 43.5      \\
\#2                  &                      & \checkmark                       & 40.1      & 42.4     & 43.0      \\
\#3                  & \checkmark                    & \checkmark                       & 42.1      & 43.4     & 44.4     \\
\hline
\end{tabular}
\caption{Ablation study on ICD.}
\label{table:ICD}
\end{center}
\end{table}
In ICD, for the mixed images, pixels from source images learn with prototype-to-pixel alignment while pixels from target images learn with point-to-pixel alignment.
As shown in Tab. \ref{table:ICD}, we conduct ablation studies on three options: 1) In Row \#1 2D network learns mixed images and only pixels from target images are constrained via point-to-pixel alignment; 2) In Row \#2 2D network learns source images and all pixels are constrained via prototype-to-pixel alignment; 3) In Row \#3 2D network learns mixed images and both source pixels and target pixels are constrained via prototype-to-pixel alignment and point-to-pixel alignment respectively. 
The comparison between Row \#1 and Row \#3 shows that although generating mixed images can make pixels from source images paired with 3D data, model performance will decrease without explicit constraints on these pixels. Our prototype-to-pixel alignment provides an effective explicit constraint.
The comparison between Row \#2 and Row \#3 demonstrates directly aligning the 2D source data with the 3D target data is difficult because the gap between them is huge, and sample mixing can help address this problem.
\newline

\noindent{\bf Analyse of ICG.}
\begin{table}[ht]
\begin{center}
\begin{tabular}{cccccc}
\hline
\multicolumn{1}{l}{} & \multirow{2}{*}{$\hat{Y}_{2D}$} & \multirow{2}{*}{$\hat{Y}_{3D}$} & \multicolumn{3}{c}{GTA5/Sem.KITTI} \\ \cline{4-6} 
\multicolumn{1}{l}{} &                      &                        & 2D     & 3D    & avg    \\
\hline
\#1                  & \checkmark                     &                        & 42.2      & 39.4     & 43.3      \\
\#2                  &                      & \checkmark                       & 41.3      & 43.5     & 44.8      \\
\#3                  & \checkmark                    & \checkmark                       & 42.2      & 44.5     & 45.6     \\
\hline
\end{tabular}
\caption{Ablation on the supervision of mixed LiDAR in ICG.
}
\label{table:ICG}
\end{center}
\end{table}
As shown in Tab. \ref{table:ICG}, we study three types of pseudo-labels for mixed LiDAR in ICG. 1) In Row \#1 3D network learns with the pseudo-labels output by 2D network; 2) In Row \#2 3D network learns with the pseudo-labels output by 3D EMA teacher model; 3) In Row \#3 3D network learns with the pseudo-labels from both 2D network and 3D EMA teacher model.
Contrary to Row \#1 which only uses the output of 2D network as supervision, Row \#3 introduces more stable pseudo-labels from 3D EMA teacher model and achieves better performance. 
However, model performance declines when relying on 3D EMA teacher model alone in Row \#2, proving that pseudo-labels from 2D model are also essential. Therefore, the two types of pseudo-labels used together can help 3D network achieve better segmentation performance. 
\newline
\noindent{\bf Comparison with multi-modal setting.}
Here, we further study the proposed setting and the developed CoMoDaL by making comparisons with the existing multi-modal learning methods~\cite{jaritz2019xmuda,peng2021sparse,10.1145/3503161.3547987}. We compare the performance of CoMoDaL in our setting (`U-M') with the performance of these methods in the multi-modal domain adaptation setting (`M-M') in Tab.~\ref{tabel:MM2SM}. 
We provide the scores reported in the papers.
According to Tab.~\ref{tabel:MM2SM}, in the absence of 3D data in source domain, CoMoDaL achieves overwhelming performance.
Specifically, for the A2D2-to-SemanticKITTI case, CoMoDaL outperforms xMUDA~\cite{jaritz2019xmuda}, DsCML~\cite{peng2021sparse} and SSE-xMUDA~\cite{10.1145/3503161.3547987} by a large margin on 2D and is also comparable to them on 3D. For the nuScenes Day-to-Night case, compared with xMUDA, DsCML and SSE-xMUDA, CoMoDaL still gets similar performance on 2D, but drops on 3D. This is because LiDAR is an active sensor that emits laser beams, which are mostly invariant to lighting conditions. Hence, 3D source data is critical to the model's performance in target domain in nuScenes Day-to-Night case. 
In contrast, the LiDAR domain gap is huge in the A2D2-to-SemanticKITTI case, CoMoDaL achieves fine performance without source LiDAR. 

\begin{table}\small
\begin{center}

\begin{tabular}{cccccccc}
\hline
\multirow{2}{*}{Method} & \multirow{2}{*}{Setting} & \multicolumn{3}{c}{A2D2/Sem.KITTI} & \multicolumn{3}{c}{nuScenes Day/Night} \\\cline{3-8} 
                        &                          & 2D     & 3D     & Avg    & 2D     & 3D     & Avg \\
\hline
xMUDA                   & M-M                      & 36.8      & 43.3      & 42.9      & 46.2      & 44.2      & 50.0\\
xMUDA+PL                & M-M                      & 43.7      & 48.5      & 49.1      & 47.1      & 46.7      & 50.8\\
DsCML                   & M-M                      & 46.3      & 50.7      & 51.0      & 49.5      & 48.2      & 52.7\\
DsCML+PL                & M-M                      & 46.8      & 51.8      & 52.4     & 50.1      & 48.7      & 53.0 \\
SSE-xMUDA                & M-M                      & 44.5      & 46.6      & 48.4      & 52.2      & 46.3      & 56.5\\
SSE-xMUDA+PL                   & M-M                      & 45.1      & 50.7      & 52.1      & 52.6      & 47.0      & 56.7\\
\hline
CoMoDaL                    & U-M                      & 48.6      & 48.2      & 50.2      & 47.5      & 37.0      & 37.9\\
CoMoDaL+PL                 & U-M                      & 49.5      & 49.4      & 51.0     & 47.6      & 37.1     & 37.4\\
\hline
\end{tabular}
\caption{Comparison results of methods in different settings. `M-M' means multi-modal domain adaptation, having multi-modal data in both domains; `U-M' is our setting, which has uni-modal data in source domain and multi-modal data in target domain.
}
\label{tabel:MM2SM}
\end{center}
\end{table}
\vspace{4.5pt}

\section{Conclusion}
    This paper aims at learning a 3DLSS model in a new and challenging setting where unlabeled image-LiDAR data in target domain and labeled 2D images in source domain are available.
To achieve it, we proposed an effective framework with a cross-modal and cross-domain learning strategy (CoMoDaL) to fully associate the data in different domains and modalities.
\method is mainly achieved through inter-modal cross-domain distillation (ICD) and intra-domain cross-modal guidance (ICG) with mixed samples in both 2D and 3D modalities.  
We conducted experiments on several datasets. The results show that our method is able to achieve meaningful segmentation performance in the proposed setting. 

\begin{acks}
 This work was supported in part by the NSFC Grant 62076101, in part by Guangdong Basic and Applied Basic Research Foundation under Grant 2023A1515010007, in part by the Guangdong Provincial Key Laboratory of Human Digital Twin under Grant 2022B1212010004, and in part by CAAI-Huawei MindSpore Open Fund.

\end{acks}

\printbibliography

@String{Computing = "Computing" }

@String{Computer = "{IEEE} Computer" }

@String{Springer = "Springer-Verlag" }

@inproceedings{vu2019advent,
  title={Advent: Adversarial entropy minimization for domain adaptation in semantic segmentation},
  author={Vu, Tuan-Hung and Jain, Himalaya and Bucher, Maxime and Cord, Matthieu and P{\'e}rez, Patrick},
  booktitle={Proceedings of the IEEE/CVF Conference on Computer Vision and Pattern Recognition},
  pages={2517--2526},
  year={2019}
}

@inproceedings{jaritz2019xmuda,
	title={{xMUDA}: Cross-Modal Unsupervised Domain Adaptation for {3D} Semantic Segmentation},
	author={Jaritz, Maximilian and Vu, Tuan-Hung and de Charette, Raoul and Wirbel, Emilie and P{\'e}rez, Patrick},
	booktitle={CVPR},
	year={2020}
}

@inproceedings{peng2021sparse,
  title={Sparse-to-dense Feature Matching: Intra and Inter domain Cross-modal Learning in Domain Adaptation for 3D Semantic Segmentation},
  author={Peng, Duo and Lei, Yinjie and Li, Wen and Zhang, Pingping and Guo, Yulan},
  booktitle={Proceedings of the International Conference on Computer Vision (ICCV)},
  year={2021},
  publisher={IEEE}
}

@InProceedings{Cordts_2016_CVPR,
author = {Cordts, Marius and Omran, Mohamed and Ramos, Sebastian and Rehfeld, Timo and Enzweiler, Markus and Benenson, Rodrigo and Franke, Uwe and Roth, Stefan and Schiele, Bernt},
title = {The Cityscapes Dataset for Semantic Urban Scene Understanding},
booktitle = {Proceedings of the IEEE Conference on Computer Vision and Pattern Recognition (CVPR)},
month = {June},
year = {2016}
}

@InProceedings{Richter2016eccv,
author="Richter, Stephan R.
and Vineet, Vibhav
and Roth, Stefan
and Koltun, Vladlen",
editor="Leibe, Bastian
and Matas, Jiri
and Sebe, Nicu
and Welling, Max",
title="Playing for Data: Ground Truth from Computer Games",
booktitle="Computer Vision -- ECCV 2016",
year="2016",
publisher="Springer International Publishing",
address="Cham",
pages="102--118",
isbn="978-3-319-46475-6"
}

@INPROCEEDINGS{wu2018squeezeseg,
  author={Wu, Bichen and Wan, Alvin and Yue, Xiangyu and Keutzer, Kurt},
  booktitle={2018 IEEE International Conference on Robotics and Automation (ICRA)}, 
  title={SqueezeSeg: Convolutional Neural Nets with Recurrent CRF for Real-Time Road-Object Segmentation from 3D LiDAR Point Cloud}, 
  year={2018},
  volume={},
  number={},
  pages={1887-1893},
  doi={10.1109/ICRA.2018.8462926}}

@inproceedings{wu2019squeezesegv2,
  title={Squeezesegv2: Improved model structure and unsupervised domain adaptation for road-object segmentation from a lidar point cloud},
  author={Wu, Bichen and Zhou, Xuanyu and Zhao, Sicheng and Yue, Xiangyu and Keutzer, Kurt},
  booktitle={2019 International Conference on Robotics and Automation (ICRA)},
  pages={4376--4382},
  year={2019},
  organization={IEEE}
}

@InProceedings{xu2021squeezesegv3,
author="Xu, Chenfeng
and Wu, Bichen
and Wang, Zining
and Zhan, Wei
and Vajda, Peter
and Keutzer, Kurt
and Tomizuka, Masayoshi",
editor="Vedaldi, Andrea
and Bischof, Horst
and Brox, Thomas
and Frahm, Jan-Michael",
title="SqueezeSegV3: Spatially-Adaptive Convolution for Efficient Point-Cloud Segmentation",
booktitle="Computer Vision -- ECCV 2020",
year="2020",
publisher="Springer International Publishing",
address="Cham",
pages="1--19",
isbn="978-3-030-58604-1"
}

@article{riegler2017octnet,
author    = {Gernot Riegler and
               Ali Osman Ulusoy and
               Andreas Geiger},
title     = {OctNet: Learning Deep 3D Representations at High Resolutions},
booktitle = {2017 {IEEE} Conference on Computer Vision and Pattern Recognition,
           {CVPR} 2017, Honolulu, HI, USA, July 21-26, 2017},
pages     = {6620--6629},
publisher = {{IEEE} Computer Society},
year      = {2017},
url       = {https://doi.org/10.1109/CVPR.2017.701},
}

@article{graham2015sparse,
author    = {Ben Graham},
editor    = {Xianghua Xie and
           Mark W. Jones and
           Gary K. L. Tam},
title     = {Sparse 3D convolutional neural networks},
booktitle = {Proceedings of the British Machine Vision Conference 2015, {BMVC}
           2015, Swansea, UK, September 7-10, 2015},
pages     = {150.1--150.9},
publisher = {{BMVA} Press},
year      = {2015},
url       = {https://doi.org/10.5244/C.29.150},
}

@article{su2018splatnet,
author    = {Hang Su and
               Varun Jampani and
               Deqing Sun and
               Subhransu Maji and
               Evangelos Kalogerakis and
               Ming{-}Hsuan Yang and
               Jan Kautz},
  title     = {SPLATNet: Sparse Lattice Networks for Point Cloud Processing},
  booktitle = {2018 {IEEE} Conference on Computer Vision and Pattern Recognition,
               {CVPR} 2018, Salt Lake City, UT, USA, June 18-22, 2018},
  pages     = {2530--2539},
  publisher = {Computer Vision Foundation / {IEEE} Computer Society},
  year      = {2018},
  url       = {http://openaccess.thecvf.com/content\_cvpr\_2018/html/Su\_SPLATNet\_Sparse\_Lattice\_CVPR\_2018\_paper.html},
}

@article{qi2017pointnet,
author    = {Charles Ruizhongtai Qi and
               Hao Su and
               Kaichun Mo and
               Leonidas J. Guibas},
  title     = {PointNet: Deep Learning on Point Sets for 3D Classification and Segmentation},
  booktitle = {2017 {IEEE} Conference on Computer Vision and Pattern Recognition,
               {CVPR} 2017, Honolulu, HI, USA, July 21-26, 2017},
  pages     = {77--85},
  publisher = {{IEEE} Computer Society},
  year      = {2017},
  url       = {https://doi.org/10.1109/CVPR.2017.16},
}

@article{qi2017pointnet++,
 author    = {Charles Ruizhongtai Qi and
               Li Yi and
               Hao Su and
               Leonidas J. Guibas},
  editor    = {Isabelle Guyon and
               Ulrike von Luxburg and
               Samy Bengio and
               Hanna M. Wallach and
               Rob Fergus and
               S. V. N. Vishwanathan and
               Roman Garnett},
  title     = {PointNet++: Deep Hierarchical Feature Learning on Point Sets in a
               Metric Space},
  booktitle = {Advances in Neural Information Processing Systems 30: Annual Conference
               on Neural Information Processing Systems 2017, December 4-9, 2017,
               Long Beach, CA, {USA}},
  pages     = {5099--5108},
  year      = {2017},
  url       = {https://proceedings.neurips.cc/paper/2017/hash/d8bf84be3800d12f74d8b05e9b89836f-Abstract.html},
}

@article{wang2019dynamic,
 author    = {Yue Wang and
               Yongbin Sun and
               Ziwei Liu and
               Sanjay E. Sarma and
               Michael M. Bronstein and
               Justin M. Solomon},
  title     = {Dynamic Graph {CNN} for Learning on Point Clouds},
  journal   = {{ACM} Trans. Graph.},
  volume    = {38},
  number    = {5},
  pages     = {146:1--146:12},
  year      = {2019},
  url       = {https://doi.org/10.1145/3326362},
}

@article{thomas2019kpconv,
      title={KPConv: Flexible and Deformable Convolution for Point Clouds}, 
      author={Hugues Thomas and Charles R. Qi and Jean-Emmanuel Deschaud and Beatriz Marcotegui and François Goulette and Leonidas J. Guibas},
      year={2019},
      eprint={1904.08889},
      archivePrefix={arXiv},
      primaryClass={cs.CV}
}

@article{zhou2020cylinder3d,
      title={Cylinder3D: An Effective 3D Framework for Driving-scene LiDAR Semantic Segmentation}, 
      author={Hui Zhou and Xinge Zhu and Xiao Song and Yuexin Ma and Zhe Wang and Hongsheng Li and Dahua Lin},
      year={2020},
      eprint={2008.01550},
      archivePrefix={arXiv},
      primaryClass={cs.CV}
}

@article{3DSemanticSegmentationWithSubmanifoldSparseConvNet,
  title={3D Semantic Segmentation with Submanifold Sparse Convolutional Networks},
  author={Graham, Benjamin and Engelcke, Martin and van der Maaten, Laurens},
  journal={CVPR},
  year={2018}
}

@inproceedings{yi2021complete,
  title={Complete \& label: A domain adaptation approach to semantic segmentation of lidar point clouds},
  author={Yi, Li and Gong, Boqing and Funkhouser, Thomas},
  booktitle={Proceedings of the IEEE/CVF Conference on Computer Vision and Pattern Recognition},
  pages={15363--15373},
  year={2021}
}

@article{ronneberger2015unet,
author    = {Olaf Ronneberger and
               Philipp Fischer and
               Thomas Brox},
  editor    = {Nassir Navab and
               Joachim Hornegger and
               William M. Wells III and
               Alejandro F. Frangi},
  title     = {U-Net: Convolutional Networks for Biomedical Image Segmentation},
  booktitle = {Medical Image Computing and Computer-Assisted Intervention - {MICCAI}
               2015 - 18th International Conference Munich, Germany, October 5 -
               9, 2015, Proceedings, Part {III}},
  series    = {Lecture Notes in Computer Science},
  volume    = {9351},
  pages     = {234--241},
  publisher = {Springer},
  year      = {2015},
  url       = {https://doi.org/10.1007/978-3-319-24574-4\_28},
}

@article{7780459,
author    = {Kaiming He and
               Xiangyu Zhang and
               Shaoqing Ren and
               Jian Sun},
  title     = {Deep Residual Learning for Image Recognition},
  booktitle = {2016 {IEEE} Conference on Computer Vision and Pattern Recognition,
               {CVPR} 2016, Las Vegas, NV, USA, June 27-30, 2016},
  pages     = {770--778},
  publisher = {{IEEE} Computer Society},
  year      = {2016},
  url       = {https://doi.org/10.1109/CVPR.2016.90},
}

@article{graham20173d,
author    = {Benjamin Graham and
               Martin Engelcke and
               Laurens van der Maaten},
  title     = {3D Semantic Segmentation With Submanifold Sparse Convolutional Networks},
  booktitle = {2018 {IEEE} Conference on Computer Vision and Pattern Recognition,
               {CVPR} 2018, Salt Lake City, UT, USA, June 18-22, 2018},
  pages     = {9224--9232},
  publisher = {Computer Vision Foundation / {IEEE} Computer Society},
  year      = {2018},
  url       = {http://openaccess.thecvf.com/content\_cvpr\_2018/html/Graham\_3D\_Semantic\_Segmentation\_CVPR\_2018\_paper.html},
}

@article{chen2017rethinking,
author    = {Liang{-}Chieh Chen and
               George Papandreou and
               Florian Schroff and
               Hartwig Adam},
  title     = {Rethinking Atrous Convolution for Semantic Image Segmentation},
  journal   = {CoRR},
  volume    = {abs/1706.05587},
  year      = {2017},
  url       = {http://arxiv.org/abs/1706.05587},
}

@article{kingma2017adam,
author    = {Diederik P. Kingma and
               Jimmy Ba},
  editor    = {Yoshua Bengio and
               Yann LeCun},
  title     = {Adam: {A} Method for Stochastic Optimization},
  booktitle = {3rd International Conference on Learning Representations, {ICLR} 2015,
               San Diego, CA, USA, May 7-9, 2015, Conference Track Proceedings},
  year      = {2015},
  url       = {http://arxiv.org/abs/1412.6980},
}

@article{geyer2020a2d2,

title={{A2D2: Audi Autonomous Driving Dataset}},

author={Jakob Geyer and Yohannes Kassahun and Mentar Mahmudi and Xavier Ricou and Rupesh Durgesh and Andrew S. Chung and Lorenz Hauswald and Viet Hoang Pham and Maximilian M{\"u}hlegg and Sebastian Dorn and Tiffany Fernandez and Martin J{\"a}nicke and Sudesh Mirashi and Chiragkumar Savani and Martin Sturm and Oleksandr Vorobiov and Martin Oelker and Sebastian Garreis and Peter Schuberth},

year={2020},

eprint={2004.06320},

archivePrefix={arXiv},

primaryClass={cs.CV},

url = {https://www.a2d2.audi}

}

@article{behley2019iccv,
author    = {Jens Behley and
               Martin Garbade and
               Andres Milioto and
               Jan Quenzel and
               Sven Behnke and
               Cyrill Stachniss and
               J{\"{u}}rgen Gall},
  title     = {SemanticKITTI: {A} Dataset for Semantic Scene Understanding of LiDAR
               Sequences},
  booktitle = {2019 {IEEE/CVF} International Conference on Computer Vision, {ICCV}
               2019, Seoul, Korea (South), October 27 - November 2, 2019},
  pages     = {9296--9306},
  publisher = {{IEEE}},
  year      = {2019},
  url       = {https://doi.org/10.1109/ICCV.2019.00939},
}

@InProceedings{Yi_2021_CVPR,
    author    = {Yi, Li and Gong, Boqing and Funkhouser, Thomas},
    title     = {Complete \& Label: A Domain Adaptation Approach to Semantic Segmentation of LiDAR Point Clouds},
    booktitle = {Proceedings of the IEEE/CVF Conference on Computer Vision and Pattern Recognition (CVPR)},
    month     = {June},
    year      = {2021},
    pages     = {15363-15373}
}

@InProceedings{10.1007/978-3-031-19827-4_34,
author="Saltori, Cristiano
and Galasso, Fabio
and Fiameni, Giuseppe
and Sebe, Nicu
and Ricci, Elisa
and Poiesi, Fabio",
editor="Avidan, Shai
and Brostow, Gabriel
and Ciss{\'e}, Moustapha
and Farinella, Giovanni Maria
and Hassner, Tal",
title="CoSMix: Compositional Semantic Mix for Domain Adaptation in 3D LiDAR Segmentation",
booktitle="Computer Vision -- ECCV 2022",
year="2022",
publisher="Springer Nature Switzerland",
address="Cham",
pages="586--602",
isbn="978-3-031-19827-4"
}

@InProceedings{10.1007/978-3-031-19827-4_33,
author="Saltori, Cristiano
and Krivosheev, Evgeny
and Lathuili{\'e}re, St{\'e}phane
and Sebe, Nicu
and Galasso, Fabio
and Fiameni, Giuseppe
and Ricci, Elisa
and Poiesi, Fabio",
editor="Avidan, Shai
and Brostow, Gabriel
and Ciss{\'e}, Moustapha
and Farinella, Giovanni Maria
and Hassner, Tal",
title="GIPSO: Geometrically Informed Propagation for Online Adaptation in 3D LiDAR Segmentation",
booktitle="Computer Vision -- ECCV 2022",
year="2022",
publisher="Springer Nature Switzerland",
address="Cham",
pages="567--585",
isbn="978-3-031-19827-4"
}

@inproceedings{lee2013pseudo,
  title={Pseudo-label: The simple and efficient semi-supervised learning method for deep neural networks},
  author={Lee, Dong-Hyun and others},
  booktitle={Workshop on challenges in representation learning, ICML},
  volume={3},
  number={2},
  pages={896},
  year={2013}
}

@INPROCEEDINGS{9423032,
  author={Tranheden, Wilhelm and Olsson, Viktor and Pinto, Juliano and Svensson, Lennart},
  booktitle={2021 IEEE Winter Conference on Applications of Computer Vision (WACV)}, 
  title={DACS: Domain Adaptation via Cross-domain Mixed Sampling}, 
  year={2021},
  volume={},
  number={},
  pages={1378-1388},
  doi={10.1109/WACV48630.2021.00142}}

@INPROCEEDINGS{9710814,
  author={Liu, Yahao and Deng, Jinhong and Gao, Xinchen and Li, Wen and Duan, Lixin},
  booktitle={2021 IEEE/CVF International Conference on Computer Vision (ICCV)}, 
  title={BAPA-Net: Boundary Adaptation and Prototype Alignment for Cross-domain Semantic Segmentation}, 
  year={2021},
  volume={},
  number={},
  pages={8781-8791},
  doi={10.1109/ICCV48922.2021.00868}}

@INPROCEEDINGS{9879251,
  author={Huo, Xinyue and Xie, Lingxi and Hu, Hengtong and Zhou, Wengang and Li, Houqiang and Tian, Qi},
  booktitle={2022 IEEE/CVF Conference on Computer Vision and Pattern Recognition (CVPR)}, 
  title={Domain-Agnostic Prior for Transfer Semantic Segmentation}, 
  year={2022},
  volume={},
  number={},
  pages={7065-7075},
  doi={10.1109/CVPR52688.2022.00694}}

@InProceedings{10.1007/978-3-031-20056-4_3,
author="Lee, Geon
and Eom, Chanho
and Lee, Wonkyung
and Park, Hyekang
and Ham, Bumsub",
editor="Avidan, Shai
and Brostow, Gabriel
and Ciss{\'e}, Moustapha
and Farinella, Giovanni Maria
and Hassner, Tal",
title="Bi-directional Contrastive Learning for Domain Adaptive Semantic Segmentation",
booktitle="Computer Vision -- ECCV 2022",
year="2022",
publisher="Springer Nature Switzerland",
address="Cham",
pages="38--55",
isbn="978-3-031-20056-4"
}

@InProceedings{Liu_2021_CVPR,
    author    = {Liu, Zhengzhe and Qi, Xiaojuan and Fu, Chi-Wing},
    title     = {3D-to-2D Distillation for Indoor Scene Parsing},
    booktitle = {Proceedings of the IEEE/CVF Conference on Computer Vision and Pattern Recognition (CVPR)},
    month     = {June},
    year      = {2021},
    pages     = {4464-4474}
}

@misc{olsson2020classmix,
      title={ClassMix: Segmentation-Based Data Augmentation for Semi-Supervised Learning}, 
      author={Viktor Olsson and Wilhelm Tranheden and Juliano Pinto and Lennart Svensson},
      year={2020},
      eprint={2007.07936},
      archivePrefix={arXiv},
      primaryClass={cs.CV}
}

@InProceedings{Hoyer_2021_CVPR,
    author    = {Hoyer, Lukas and Dai, Dengxin and Chen, Yuhua and Koring, Adrian and Saha, Suman and Van Gool, Luc},
    title     = {Three Ways To Improve Semantic Segmentation With Self-Supervised Depth Estimation},
    booktitle = {Proceedings of the IEEE/CVF Conference on Computer Vision and Pattern Recognition (CVPR)},
    month     = {June},
    year      = {2021},
    pages     = {11130-11140}
}

@misc{chen2020pointmixup,
      title={PointMixup: Augmentation for Point Clouds}, 
      author={Yunlu Chen and Vincent Tao Hu and Efstratios Gavves and Thomas Mensink and Pascal Mettes and Pengwan Yang and Cees G. M. Snoek},
      year={2020},
      eprint={2008.06374},
      archivePrefix={arXiv},
      primaryClass={cs.CV}
}

@misc{kong2023lasermix,
      title={LaserMix for Semi-Supervised LiDAR Semantic Segmentation}, 
      author={Lingdong Kong and Jiawei Ren and Liang Pan and Ziwei Liu},
      year={2023},
      eprint={2207.00026},
      archivePrefix={arXiv},
      primaryClass={cs.CV}
}

@misc{nekrasov2021mix3d,
      title={Mix3D: Out-of-Context Data Augmentation for 3D Scenes}, 
      author={Alexey Nekrasov and Jonas Schult and Or Litany and Bastian Leibe and Francis Engelmann},
      year={2021},
      eprint={2110.02210},
      archivePrefix={arXiv},
      primaryClass={cs.CV}
}

@inproceedings{NIPS2017_68053af2,
 author = {Tarvainen, Antti and Valpola, Harri},
 booktitle = {Advances in Neural Information Processing Systems},
 editor = {I. Guyon and U. Von Luxburg and S. Bengio and H. Wallach and R. Fergus and S. Vishwanathan and R. Garnett},
 pages = {},
 publisher = {Curran Associates, Inc.},
 title = {Mean teachers are better role models: Weight-averaged consistency targets improve semi-supervised deep learning results},
 url = {https://proceedings.neurips.cc/paper_files/paper/2017/file/68053af2923e00204c3ca7c6a3150cf7-Paper.pdf},
 volume = {30},
 year = {2017}
}

@article{nuscenes2019,
  title={nuScenes: A multimodal dataset for autonomous driving},
  author={Holger Caesar and Varun Bankiti and Alex H. Lang and Sourabh Vora and 
          Venice Erin Liong and Qiang Xu and Anush Krishnan and Yu Pan and 
          Giancarlo Baldan and Oscar Beijbom},
  journal={arXiv preprint arXiv:1903.11027},
  year={2019}
}

@misc{ding2022doda,
      title={DODA: Data-oriented Sim-to-Real Domain Adaptation for 3D Semantic Segmentation}, 
      author={Runyu Ding and Jihan Yang and Li Jiang and Xiaojuan Qi},
      year={2022},
      eprint={2204.01599},
      archivePrefix={arXiv},
      primaryClass={cs.CV}
}

@INPROCEEDINGS{9341508,
  author={Langer, Ferdinand and Milioto, Andres and Haag, Alexandre and Behley, Jens and Stachniss, Cyrill},
  booktitle={2020 IEEE/RSJ International Conference on Intelligent Robots and Systems (IROS)}, 
  title={Domain Transfer for Semantic Segmentation of LiDAR Data using Deep Neural Networks}, 
  year={2020},
  volume={},
  number={},
  pages={8263-8270},
  doi={10.1109/IROS45743.2020.9341508}}

@INPROCEEDINGS{9008296,
  author={Yun, Sangdoo and Han, Dongyoon and Chun, Sanghyuk and Oh, Seong Joon and Yoo, Youngjoon and Choe, Junsuk},
  booktitle={2019 IEEE/CVF International Conference on Computer Vision (ICCV)}, 
  title={CutMix: Regularization Strategy to Train Strong Classifiers With Localizable Features}, 
  year={2019},
  volume={},
  number={},
  pages={6022-6031},
  doi={10.1109/ICCV.2019.00612}}

@inproceedings{10.1145/3503161.3547987,
author = {Zhang, Yachao and Li, Miaoyu and Xie, Yuan and Li, Cuihua and Wang, Cong and Zhang, Zhizhong and Qu, Yanyun},
title = {Self-Supervised Exclusive Learning for 3D Segmentation with Cross-Modal Unsupervised Domain Adaptation},
year = {2022},
isbn = {9781450392037},
publisher = {Association for Computing Machinery},
address = {New York, NY, USA},
url = {https://doi.org/10.1145/3503161.3547987},
doi = {10.1145/3503161.3547987},
booktitle = {Proceedings of the 30th ACM International Conference on Multimedia},
pages = {3338–3346},
numpages = {9},
location = {Lisboa, Portugal},
series = {MM '22}
}

\appendix
\section{Additional illustration and experimental results}
\subsection{Further ablation study on ICD module}
According to the results in Tab. \ref{table:pixels}, for mixed images, it is clear to see when applying prototype-to-pixel alignment for all source pixels, the model performance declines. When utilizing 3D-2D projection of target samples to select source pixels, the model performance recovers. However, model performance declines again when randomly selecting pixels based on the number of projection points to optimize. We guess it might be because both random selection and all selection cause different 2D-3D correspondence patterns regarding the target 2D and source 2D.

We also explore the effect of different kinds of masks. The class-level mask is widely used in 2D semantic segmentation task~\cite{9423032,olsson2020classmix} to replace the region-level mask when employing CutMix strategy, because it can generate mixed images that better respect semantic boundaries.
However, results in Tab. \ref{table:mask} show the class-level mask and region-level mask are comparable. In fact, we have shown in Tab. \ref{table:pixels} that only sparse source pixels in mixed images should be constrained, which determines that mixed images cannot learn the boundary information effectively brought by class-level mask with dense supervision like prior works~\cite{9423032,olsson2020classmix} in our setting.
\begin{table}[b]\small
\hfill
\begin{minipage}[t]{0.48\linewidth}
\vspace{0pt}
\centering{
\begin{tabular}{cccc}
\hline
\multirow{2}{*}{Method} & \multicolumn{3}{c}{GTA5/Sem.KITTI} \\ \cline{2-4} 
                        & 2D        & 3D        & Avg        \\
\hline
all                   & 39.1         & 39.4         & 41.1         \\
random                  & 40.3         & 40.8         & 42.5         \\
projection               & 42.1         & 43.4         & 44.4         \\
\hline
\end{tabular}}
\caption{Comparison on the number of source pixels aligned with 3D prototypes in the mixed images}
\label{table:pixels}
\end{minipage}
\hspace{.1em}
\begin{minipage}[t]{0.48\linewidth}
\vspace{0pt}
\centering{
\begin{tabular}{cccc}
\hline
\multirow{2}{*}{Method} & \multicolumn{3}{c}{GTA5/Sem.KITTI} \\ \cline{2-4} 
                        & 2D        & 3D        & Avg        \\
\hline
class-level  & 41.6  & 43.7  & 44.3   \\
region-level & 42.1  & 43.4  & 44.4  \\
\hline
\end{tabular}}
\vspace{10pt}
\caption{Comparison of the binary mask used to generate mixed images.}
\label{table:mask}
\end{minipage}
\end{table}
\subsection{Illustration of modal interaction for different input}
The comparison of modal interaction for target image and mixed image can be viewed in Fig. \ref{fig:method}. The main difference between them is that the mixed image
introduces source pixels, so additional prototype-to-pixel correspondence is required.

\begin{figure}[b]
  \centering
  \includegraphics[width=\linewidth]{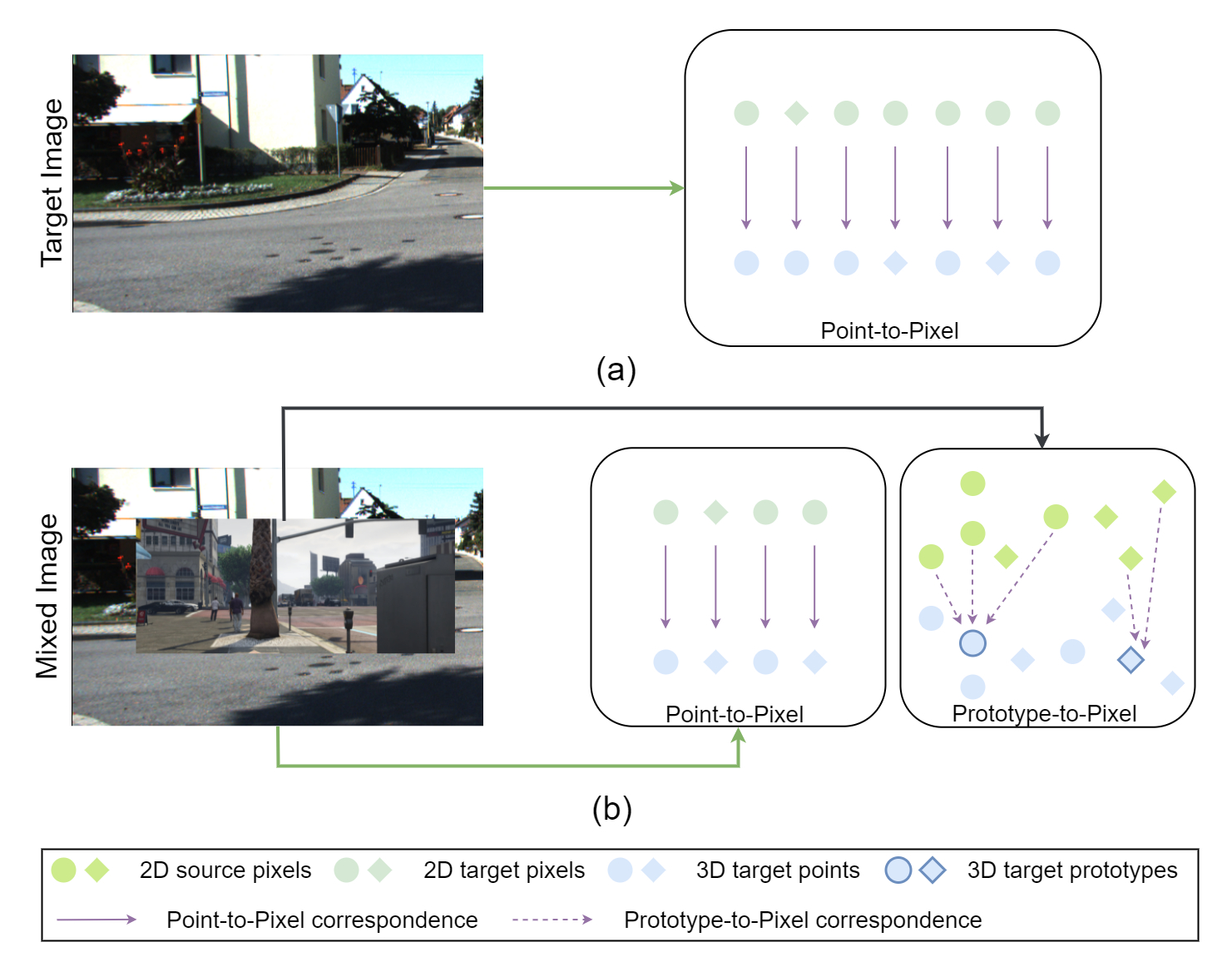} 
      \caption{Comparison of modal interaction for different input. (a) For the target image, 2D predictions are aligned with 3D predictions via point-to-pixel correspondence; (b) For the mixed image, predictions of source pixels are aligned with 3D prototypes via prototype-to-pixel correspondence while predictions of target pixels are aligned with 3D predictions via point-to-pixel correspondence.} 
  \label{fig:method} 
\end{figure}
\end{document}